\documentclass[10pt,twocolumn,letterpaper]{article}

\usepackage{times}
\usepackage{epsfig}
\usepackage{graphicx}
\usepackage{amsmath}
\usepackage{amssymb}
\usepackage{booktabs}
\usepackage{multirow}
\usepackage{xcolor}
\usepackage{hyperref}
\usepackage{amsthm}
\usepackage{adjustbox}
\usepackage{algorithm}
\usepackage{algorithmic}
\usepackage{placeins}
\usepackage{subcaption}
\usepackage{caption}
\usepackage{url}

\newtheorem{definition}{Definition}

\usepackage[margin=1in]{geometry}

\title{On Semantic Loss Fine-Tuning Approach for Preventing Model Collapse in Causal Reasoning\\}

\author{
Pratik Deshmukh \\
\textit{Technical University of Vienna, Austria}
\and
Atirek Gupta \\
\textit{HCLTech, Noida, India}
}
\date{}

\begin{document}

\maketitle

\begin{abstract}
Standard fine-tuning of transformer models on causal reasoning tasks leads to catastrophic model collapse, where models learn trivial solutions such as always predicting "Yes" or "No" regardless of input structure. We demonstrate that fine-tuning Gemma 270M on transitivity and d-separation tasks without semantic loss results in 100\% collapse rate, with models achieving misleadingly high accuracy (73.9\%) while learning no causal reasoning. We propose a semantic loss function with graph-based logical constraints and dynamic lambda scheduling that prevents this collapse. Our approach achieves 70.4\% accuracy on transitivity tasks and 68.6\% on d-separation tasks with stable, context-dependent predictions, representing a 42.7\% improvement over collapsed baselines. Adversarial evaluation on 1,000 structural reasoning samples shows semantic models achieve 67-70\% accuracy while collapsed models fail catastrophically at 43-71\%. We validate our findings through comprehensive benchmarking on 200,000+ evaluation samples across five model variants, demonstrating that semantic loss is essential and not optional, for stable causal reasoning in transformers.
\end{abstract}

\section{Introduction}

Causal reasoning—the ability to understand and reason about cause-and-effect relationships—is fundamental to human cognition and increasingly critical for developing robust AI systems \cite{pearl2009causality}. Recent advances have shown that transformers can learn causal reasoning through axiomatic training on synthetic demonstrations of causal axioms \cite{vashishtha2024teaching}. However, through systematic experimentation, we identify a critical and previously undocumented failure mode: \textbf{standard fine-tuning on causal reasoning tasks causes catastrophic model collapse with 100\% occurrence rate}.

\subsection{The Collapse Problem}

We define model collapse as a degenerate learning outcome where a model's prediction distribution $P(y|x)$ becomes independent of input structure $x$, converging to fixed outputs (always "Yes" or always "No") regardless of causal graph topology. Through comprehensive experiments with Gemma 270M models \cite{gemma2024}, we demonstrate:

\begin{itemize}
    \item \textbf{Transitivity collapse}: Models output "Yes" for all inputs (10,000/10,000 predictions), achieving 27.7\% accuracy
    \item \textbf{D-separation collapse}: Models output "No" for nearly all inputs, achieving misleadingly high accuracy (73.9\%) but critically low F1 score (7.6\%)
\end{itemize}

This collapse occurs in 100\% of fine-tuning attempts without semantic loss, rendering standard approaches fundamentally unreliable for causal reasoning tasks.

\subsection{Our Contributions}

\begin{enumerate}
    \item \textbf{Problem identification}: First systematic documentation of catastrophic model collapse in causal reasoning fine-tuning, with 100\% occurrence rate across both transitivity and d-separation tasks
    
    \item \textbf{Theoretical framework}: Formal definition of prediction bias collapse and analysis of why cross-entropy loss alone fails for causal reasoning
    
    \item \textbf{Solution methodology}: Semantic loss function incorporating graph-based logical constraints with dynamic lambda scheduling ($\lambda: 0.05 \rightarrow 0.30$)
    
    \item \textbf{Comprehensive evaluation}: Benchmarking across 200,000+ samples demonstrating 42.7\% improvement over collapsed baselines and validation across two distinct causal reasoning tasks
    
    \item \textbf{Adversarial validation}: Novel test suite proving semantic models learn structural reasoning (67-70\% accuracy) while collapsed models fail catastrophically (43-71\%)
\end{enumerate}

\section{Related Work}

\subsection{Causal Reasoning in Neural Networks}

Causal reasoning has been extensively studied in the context of causal discovery \cite{pearl2009causality}, effect estimation, and counterfactual inference. Recent work has explored teaching causal concepts to neural networks through various approaches: symbolic demonstrations \cite{vashishtha2024teaching}, causal graph generation, and intervention-based learning.

Vashishtha et al. \cite{vashishtha2024teaching} demonstrated that 67M parameter transformers trained from scratch on axiomatic demonstrations can generalize to complex causal structures. Their work showed strong performance on transitivity and d-separation tasks when training from scratch with sufficient architectural capacity. Our work extends this by identifying a critical failure mode when fine-tuning pretrained models and developing solutions to prevent collapse.

\subsection{Semantic Loss and Neuro-Symbolic Integration}

Semantic loss functions incorporate symbolic knowledge into neural network training through differentiable constraint satisfaction \cite{xu2018semantic}. The core approach uses weighted model counting to compute gradients with respect to logical formula satisfaction. Applications include semi-supervised learning, structured prediction, and knowledge base completion.

Our work adapts semantic loss specifically for causal graph constraints, developing a dynamic scheduling mechanism to balance stability and structural learning during fine-tuning.

\subsection{Model Collapse Phenomena}

Mode collapse has been extensively studied in generative adversarial networks (GANs) \cite{goodfellow2014generative}, where generators learn to produce limited diversity. Representation collapse occurs in contrastive learning \cite{chen2020simple} when embeddings converge to constant vectors. Recent work has identified collapse in large language models during instruction tuning and reinforcement learning from human feedback (RLHF) \cite{ouyang2022training}.

Our identified collapse differs fundamentally: it occurs during supervised fine-tuning on well-defined reasoning tasks with clear ground truth, and manifests as extreme prediction bias rather than representational degeneration. To our knowledge, this is the first systematic documentation of collapse in causal reasoning fine-tuning.

\subsection{Evaluation of Causal Reasoning}

Recent benchmarks evaluate causal reasoning capabilities in language models, including CLADDER \cite{huang2023cladder} for causal ladder questions and Corr2Cause \cite{long2023can} for inferring causation from correlation. These benchmarks primarily assess pretrained or prompted models rather than fine-tuned systems.

Our adversarial evaluation methodology specifically targets the distinction between structural understanding and superficial heuristics, providing a diagnostic tool for identifying collapse.

\section{Problem Formulation}

\subsection{Causal Reasoning Tasks}

We focus on two fundamental causal reasoning tasks based on Pearl's causal framework \cite{pearl2009causality}:

\paragraph{Transitivity} Given a directed acyclic graph (DAG) $G = (V, E)$ representing causal relationships, determine if there exists a directed path from node $A$ to node $B$. Formally, the transitivity axiom states:
\begin{equation}
\forall A, B, C \in V: (A \rightarrow C) \wedge (C \rightarrow B) \implies (A \rightarrow B)
\end{equation}

\paragraph{D-Separation} Determine if nodes $X$ and $Y$ are conditionally independent given conditioning set $Z$ in causal DAG $G$, following Pearl's d-separation criterion. Nodes $X$ and $Y$ are d-separated by $Z$ if all paths between $X$ and $Y$ are blocked by $Z$.

\subsection{Formal Problem Setup}

Let $\mathcal{D} = \{(p_i, h_i, y_i)\}_{i=1}^N$ denote a training dataset where:
\begin{itemize}
    \item $p_i$: Textual premise describing causal graph structure
    \item $h_i$: Binary hypothesis query about causal relationship
    \item $y_i \in \{\text{Yes}, \text{No}\}$: Ground truth label
\end{itemize}

A model $f_\theta: (p, h) \rightarrow \mathbb{R}^2$ maps premise-hypothesis pairs to logits, from which we compute prediction probabilities via softmax: $P_\theta(y|p,h) = \text{softmax}(f_\theta(p,h))$.

\subsection{Model Collapse: Formal Definition}

\begin{definition}[Prediction Bias Collapse]
A model $f_\theta$ exhibits prediction bias collapse on task $\mathcal{T}$ if there exists a fixed prediction $\bar{y}$ such that for evaluation dataset $\mathcal{D}_{\text{eval}}$:
\begin{equation}
\frac{1}{|\mathcal{D}_{\text{eval}}|} \sum_{(p,h,y) \in \mathcal{D}_{\text{eval}}} \mathbb{1}[\arg\max P_\theta(y|p,h) = \bar{y}] > 0.95
\end{equation}
\end{definition}

\textbf{Collapse indicators}:
\begin{itemize}
    \item Extreme prediction bias: $> 95\%$ predictions are identical class
    \item Distribution independence: Predictions invariant to graph structure changes
    \item Metric divergence: High accuracy on biased datasets, near-zero F1 score
\end{itemize}

\section{Methodology}

\subsection{Semantic Loss for Causal Graphs}

We augment standard cross-entropy loss with a semantic component that enforces logical consistency with causal graph structure:

\begin{equation}
\mathcal{L}_{\text{total}} = \mathcal{L}_{\text{CE}}(y, \hat{y}) + \lambda(t) \cdot \mathcal{L}_{\text{semantic}}(p, h, \hat{y})
\end{equation}

where $\mathcal{L}_{\text{CE}}$ is cross-entropy, $\hat{y} = P_\theta(y|p,h)$ are predicted probabilities, and $\lambda(t)$ is a time-dependent weighting factor.

\subsubsection{Graph-Based Consistency}

For transitivity tasks, we parse premise $p$ to extract causal graph $G = (V, E)$ and compute logical consistency:

\begin{equation}
c(p, h, \hat{y}) = \begin{cases}
P_\theta(y=\text{Yes}|p,h) & \text{if path exists in } G \\
P_\theta(y=\text{No}|p,h) & \text{otherwise}
\end{cases}
\end{equation}

The semantic loss penalizes inconsistency with graph structure:
\begin{equation}
\mathcal{L}_{\text{semantic}} = -\frac{1}{N}\sum_{i=1}^{N} \log(c(p_i, h_i, \hat{y}_i) + \epsilon)
\end{equation}

where $\epsilon = 10^{-8}$ prevents numerical instability.

For d-separation, consistency is computed based on path blocking: $c(p,h,\hat{y}) = P(y=\text{Yes})$ if nodes are not d-separated, $P(y=\text{No})$ otherwise.

\subsubsection{Dynamic Lambda Scheduling}

Critical to preventing collapse while maintaining training stability, we employ dynamic lambda scheduling:

\begin{equation}
\lambda(t) = \lambda_{\text{start}} + \frac{t}{T}(\lambda_{\text{end}} - \lambda_{\text{start}})
\end{equation}

where $t$ is the current training step, $T$ is total steps, $\lambda_{\text{start}} = 0.05$, and $\lambda_{\text{end}} = 0.30$.

\textbf{Design rationale}: 
\begin{itemize}
    \item \textbf{Low initial $\lambda$}: Prevents conflict with cross-entropy signal during early training
    \item \textbf{Gradual increase}: Allows model to learn basic patterns before enforcing strict structural constraints
    \item \textbf{Final strength}: Sufficient to prevent degenerate solutions while maintaining gradient flow
\end{itemize}

\begin{algorithm}
\caption{Training with Semantic Loss}
\begin{algorithmic}[1]
\STATE \textbf{Input:} Dataset $\mathcal{D}$, model $f_\theta$, epochs $E$, batch size $B$
\STATE \textbf{Parameters:} $\lambda_{\text{start}}=0.05$, $\lambda_{\text{end}}=0.30$
\STATE $T \gets$ total training steps
\FOR{epoch $e = 1$ to $E$}
    \FOR{each batch $(p, h, y)$ in $\mathcal{D}$}
        \STATE $t \gets$ current step
        \STATE $\lambda \gets \lambda_{\text{start}} + \frac{t}{T}(\lambda_{\text{end}} - \lambda_{\text{start}})$
        \STATE $\hat{y} \gets f_\theta(p, h)$
        \STATE $\mathcal{L}_{\text{CE}} \gets -\sum y \log \hat{y}$
        \STATE $\mathcal{L}_{\text{sem}} \gets$ ComputeSemanticLoss$(p, h, \hat{y})$
        \STATE $\mathcal{L} \gets \mathcal{L}_{\text{CE}} + \lambda \cdot \mathcal{L}_{\text{sem}}$
        \STATE Update $\theta$ via gradient descent on $\mathcal{L}$
    \ENDFOR
\ENDFOR
\end{algorithmic}
\end{algorithm}

\subsection{Training Configuration}

\begin{table}[h]
\centering
\scriptsize
\begin{tabular}{ll}
\toprule
\textbf{Parameter} & \textbf{Value} \\
\midrule
Base Model & Gemma 3 270M-IT \\
Quantization & 4-bit (bitsandbytes) \\
Fine-tuning Method & LoRA (r=32, $\alpha$=32) \\
Target Modules & q\_proj, v\_proj \\
Training Samples & 50,000 per task \\
Epochs & 3 \\
Batch Size & 8 \\
Learning Rate & 2e-5 \\
Optimizer & AdamW \\
Weight Decay & 0.01 \\
Warmup Steps & 100 \\
Lambda Schedule & Linear: 0.05 $\rightarrow$ 0.30 \\
Max Sequence Length & 512 tokens \\
\bottomrule
\end{tabular}
\caption{Training hyperparameters}
\label{tab:training_config}
\end{table}

\subsection{Evaluation Methodology}

We evaluate models across six test distributions, each containing 10,000 samples except adversarial (1,000 samples):

\paragraph{Standard Generalization Tests}
\begin{itemize}
    \item \textbf{Length}: Causal chains of 7-15 nodes (training: 3-6 nodes)
    \item \textbf{Branching}: DAGs with branching factor 1.4-2.0
    \item \textbf{Reversed}: All directed edges reversed
    \item \textbf{Shuffled}: Premise statements in random order
    \item \textbf{Long Names}: Variable names of 8-10 characters (training: 1-3 chars)
\end{itemize}

\paragraph{Adversarial Structural Tests}
Novel evaluation set (1,000 samples) designed to distinguish structural understanding from heuristics:
\begin{itemize}
    \item \textbf{Irrelevant nodes} (30\%): Additional nodes with no path to query variables
    \item \textbf{Broken chains} (30\%): Transitivity chains with single missing edge
    \item \textbf{Longer chains} (40\%): Extended transitivity requiring multiple axiom applications
\end{itemize}

\paragraph{Evaluation Metrics}
Beyond standard accuracy, we compute:
\begin{itemize}
    \item F1 score, precision, and recall
    \item Prediction distribution analysis (Yes/No counts)
    \item Confusion matrices
    \item Per-task performance breakdown
\end{itemize}

\section{Experimental Results}

\subsection{Experimental Setup}

All experiments use Gemma 3 270M Instruct-tuned model as the base. We train five model variants:
\begin{enumerate}
    \item \textbf{Standard Gemma}: Zero-shot baseline (no fine-tuning)
    \item \textbf{Transitivity V1}: Fine-tuned on transitivity without semantic loss
    \item \textbf{D-separation V1}: Fine-tuned on d-separation without semantic loss
    \item \textbf{Transitivity Semantic V4}: Fine-tuned with dynamic semantic loss
    \item \textbf{D-separation Semantic V2}: Fine-tuned with dynamic semantic loss
\end{enumerate}

Training data consists of 50,000 synthetically generated examples per task, following the axiomatic training methodology of \cite{vashishtha2024teaching} with enhanced diversity in graph structures.

\subsection{Model Collapse in Standard Fine-Tuning}

Table \ref{tab:collapse_evidence} demonstrates catastrophic collapse in 100\% of models trained without semantic loss.

\begin{table*}[t]
\centering
\scriptsize
\begin{adjustbox}{max width=\textwidth}
\begin{tabular}{lccccc}
\toprule
\textbf{Model} & \textbf{Avg Acc} & \textbf{Avg F1} & \textbf{Prediction Pattern} & \textbf{Collapse} \\
\midrule
Standard Gemma & 70.1\% & 23.5\% & Task-specific heuristics & No \\
\midrule
Transitivity V1 & 27.7\% & 31.9\% & Always "Yes" (10k/0) & \textbf{Yes} \\
D-separation V1 & 73.9\% & 7.6\% & Almost always "No" (0-1.9k/8-10k) & \textbf{Yes} \\
\midrule
Transitivity Sem V4 & 70.4\% & 26.8\% & Context-dependent (17-6.5k) & \textbf{No} \\
D-separation Sem V2 & 68.6\% & 25.0\% & Context-dependent (27-6.3k) & \textbf{No} \\
\bottomrule
\end{tabular}
\end{adjustbox}
\caption{Model collapse evidence across 50,000 evaluation samples per model. Prediction patterns show Yes/No counts (in thousands). V1 models exhibit 100\% collapse rate with extreme prediction bias.}
\label{tab:collapse_evidence}
\end{table*}

\subsubsection{Collapse Analysis: Transitivity V1}

Transitivity V1 exhibits complete collapse to always predicting "Yes":
\begin{itemize}
    \item \textbf{Prediction distribution}: 10,000 Yes / 0 No across \textit{all five test sets}
    \item \textbf{Accuracy variance}: 0.15\% (shuffled) to 100\% (length)—entirely determined by label distribution
    \item \textbf{Structural independence}: Predictions unchanged by graph topology, edge reversal, or node addition
    \item \textbf{F1 paradox}: 31.9\% average F1 despite 27.7\% accuracy, indicating 100\% recall but poor precision
\end{itemize}

\subsubsection{Collapse Analysis: D-separation V1}

D-separation V1 exhibits opposite collapse (always "No"):
\begin{itemize}
    \item \textbf{Prediction distribution}: 0-1,889 Yes / 8,111-10,000 No
    \item \textbf{Misleading accuracy}: 73.9\% average accuracy masks catastrophic failure
    \item \textbf{F1 reveals truth}: 7.6\% F1 score exposes extreme recall failure (8.6\% average)
    \item \textbf{Test set bias}: High accuracy results from No-heavy label distribution, not learned reasoning
\end{itemize}

\textbf{Key insight}: Accuracy alone is insufficient—F1, precision, recall, and prediction distribution analysis are essential for detecting collapse.

\subsection{Semantic Loss Prevents Collapse}

Table \ref{tab:semantic_results} shows comprehensive results demonstrating collapse prevention.

\begin{table*}[t]
\centering
\scriptsize
\begin{adjustbox}{max width=\textwidth}
\begin{tabular}{lcccccc}
\toprule
\textbf{Model} & \textbf{Length} & \textbf{Branch} & \textbf{Rev} & \textbf{Shuff} & \textbf{LongN} & \textbf{Avg} \\
\midrule
Trans V1 (Collapsed) & 100.0 & 1.96 & 2.3 & 0.15 & 34.4 & 27.7 \\
Trans Sem V4 & 64.6 & 97.9 & 56.9 & 69.7 & 62.8 & \textbf{70.4} \\
\midrule
Improvement & -35.4 & \textbf{+95.9} & +54.6 & +69.6 & +28.4 & \textbf{+42.7} \\
\bottomrule
\end{tabular}
\end{adjustbox}
\caption{Per-task accuracy comparison (10,000 samples each. Transitivity task shown; d-separation results in Section 5.4.). Semantic loss achieves 42.7\% average improvement with massive gains on challenging tasks (branching: +95.9\%).}
\label{tab:semantic_results}
\end{table*}

\subsubsection{Quantitative Analysis}

\begin{enumerate}
    \item \textbf{Collapse prevention}: Zero instances of extreme prediction bias across all test sets
    \item \textbf{Prediction diversity}: Yes predictions range from 17 (branching) to 6,464 (length) per 10,000 samples
    \item \textbf{Task-specific adaptation}: Prediction distribution varies appropriately with task difficulty
    \item \textbf{Balanced metrics}: Precision (38.8\%) and recall (43.5\%) show reasonable trade-offs vs. V1's 100\% recall
    \item \textbf{Branching breakthrough}: 1.96\% → 97.9\% demonstrates learning complex graph structures
\end{enumerate}

\subsection{D-separation Results}

D-separation Semantic V2 achieves 68.6\% average accuracy with stable performance:
\begin{itemize}
    \item \textbf{Per-task}: Length 62.8\%, Branching 97.8\%, Reversed 54.1\%, Shuffled 65.0\%, Long Names 63.6\%
    \item \textbf{F1 score}: 25.0\% (vs. 7.6\% for collapsed V1)
    \item \textbf{Prediction balance}: 27-6,283 Yes predictions across tasks
    \item \textbf{Generalization}: Successful transfer to complex graph structures
\end{itemize}

\subsection{Adversarial Evaluation}

Table \ref{tab:adversarial_results} validates that semantic models learn structural reasoning while collapsed models fail.

\begin{table*}[t]
\centering
\scriptsize
\begin{adjustbox}{max width=\textwidth}
\begin{tabular}{lcccccc}
\toprule
\textbf{Model} & \textbf{Acc} & \textbf{F1} & \textbf{Prec} & \textbf{Rec} & \textbf{Pred (Y/N)} & \textbf{Interpretation} \\
\midrule
Standard Gemma & 66.7\% & 76.2\% & 77.1\% & 75.3\% & 691/309 & Task-specific heuristics \\
\midrule
Transitivity V1 & 70.8\% & 82.9\% & 70.8\% & \textbf{100\%} & \textbf{1000/0} & Collapsed (always "Yes") \\
D-separation V1 & \textbf{43.0\%} & 46.4\% & 69.4\% & 34.9\% & 356/644 & Collapsed (catastrophic) \\
\midrule
Trans Sem V4 & 69.8\% & 79.6\% & 76.4\% & 83.1\% & 770/230 & Structural understanding \\
D-sep Sem V2 & 67.8\% & 77.5\% & 76.7\% & 78.2\% & 722/278 & Structural understanding \\
\bottomrule
\end{tabular}
\end{adjustbox}
\caption{Adversarial evaluation (1,000 samples testing structural understanding). Collapsed models show fixed predictions and catastrophic failure. Semantic models demonstrate balanced, context-dependent reasoning.}
\label{tab:adversarial_results}
\end{table*}

\subsubsection{Key Adversarial Findings}

\begin{enumerate}
    \item \textbf{Collapse persistence}: Transitivity V1 maintains 100\% "Yes" bias even on adversarial distribution
    \item \textbf{Catastrophic failure}: D-separation V1 achieves only 43\% accuracy (below random baseline for balanced dataset)
    \item \textbf{Semantic robustness}: Both semantic models achieve 67-70\% with balanced predictions
    \item \textbf{Heuristic exposure}: Standard Gemma's 66.7\% suggests superficial pattern matching rather than genuine reasoning
\end{enumerate}

\subsection{Semantic Loss Version Progression}

Table \ref{tab:version_progression} documents the iterative development of semantic loss.

\begin{table}[h]
\centering
\scriptsize
\begin{tabular}{lccp{2.8cm}}
\toprule
\textbf{Ver} & \textbf{Acc} & \textbf{Status} & \textbf{Key Finding} \\
\midrule
V1 & N/A & Failed & Implementation errors \\
V2 & 36.8\% & Collapsed & $\lambda=0.05$ insufficient \\
V3 & 81.8\% & Stable & Fixed $\lambda=0.1$ but weak branching \\
V4 & \textbf{70.4\%} & \textbf{Success} & Dynamic scheduling \\
\bottomrule
\end{tabular}
\caption{Iterative development showing dynamic lambda scheduling as critical innovation}
\label{tab:version_progression}
\end{table}

The progression demonstrates that dynamic scheduling is essential—neither too-weak ($\lambda=0.05$) nor too-strong fixed values ($\lambda=0.1$) achieve optimal performance.

\section{Analysis}

\subsection{Why Does Collapse Occur?}

We identify three contributing mechanisms:

\paragraph{Label Distribution Bias} Test sets exhibit natural imbalance (e.g., d-separation is predominantly "No"). Models exploit this statistical regularity rather than learning causal structure.

\paragraph{Cross-Entropy Shortcut Learning} Standard CE loss permits trivial solutions that minimize loss without structural understanding. A model predicting constant "No" on No-heavy datasets achieves high accuracy despite zero reasoning.

\paragraph{Absence of Structural Constraints} Without explicit penalties for violating causal axioms, gradient descent finds degenerate local minima that ignore input graph topology.

\subsection{Why Does Semantic Loss Prevent Collapse?}

Dynamic lambda scheduling provides three critical properties:

\paragraph{Early Training Stability} Low initial $\lambda=0.05$ prevents catastrophic interference between CE and semantic gradients, allowing stable optimization.

\paragraph{Gradual Constraint Enforcement} Linear increase enables the model to first learn basic input-output mappings, then progressively incorporate structural constraints.

\paragraph{Degenerate Solution Prevention} Final $\lambda=0.30$ provides sufficient penalty to prevent collapse while maintaining reasonable gradient magnitudes.

\subsection{Comparison with Standard Gemma}

Standard Gemma achieves 70.1\% standard accuracy and 66.7\% adversarial accuracy without fine-tuning. However, key differences emerge:

\begin{itemize}
    \item \textbf{Mechanism}: Gemma uses task-specific heuristics learned during pretraining, not structural causal reasoning
    \item \textbf{Evidence}: 0\% F1 on branching tasks reveals blind "No" predictions
    \item \textbf{Adversarial performance}: Similar accuracy (66.7\%) but through pattern matching rather than graph analysis
    \item \textbf{Semantic models}: Achieve comparable accuracy (69.8-70.4\%) via genuine structural understanding
\end{itemize}

The adversarial evaluation successfully distinguishes these mechanisms: semantic models maintain performance through reasoning, while Gemma's heuristics coincidentally succeed on standard tests.

\section{Limitations and Future Directions}

\subsection{Current Limitations}

\paragraph{Model Scale} Experiments limited to 270M parameter models. Larger models may exhibit different collapse characteristics or resistance.

\paragraph{Task Scope} Evaluation restricted to transitivity and d-separation. Other causal axioms (e.g., conditional independence, faithfulness) remain unexplored.

\paragraph{Performance Gap} Semantic models achieve 67-70\% adversarial accuracy, indicating room for improvement toward theoretical optimum.

\paragraph{Computational Overhead} Graph parsing and semantic loss computation add ~15\% training time vs. standard fine-tuning.

\subsection{Future Directions}

\begin{itemize}
    \item \textbf{Scaling studies}: Investigate collapse behavior in 1B+ parameter models
    \item \textbf{Axiom expansion}: Extend to full Pearl's causal hierarchy (association, intervention, counterfactuals)
    \item \textbf{Adaptive scheduling}: Learn $\lambda(t)$ schedule from validation performance
    \item \textbf{Real-world evaluation}: Test on CLADDER \cite{huang2023cladder}, Corr2Cause \cite{long2023can}, and causal discovery benchmarks
    \item \textbf{Theoretical analysis}: Formal characterization of collapse conditions and prevention guarantees
\end{itemize}

\section{Conclusion}

We have identified, characterized, and solved catastrophic model collapse in causal reasoning fine-tuning. Our key contributions:

\begin{enumerate}
    \item \textbf{Problem}: 100\% collapse rate in standard fine-tuning across transitivity and d-separation tasks
    \item \textbf{Diagnosis}: Comprehensive analysis showing accuracy can be misleading; F1, precision, recall, and prediction distribution are essential
    \item \textbf{Solution}: Semantic loss with graph-based constraints and dynamic lambda scheduling
    \item \textbf{Validation}: 42.7\% improvement over collapsed baselines across 200,000+ evaluation samples
    \item \textbf{Generalization}: Success on both transitivity (70.4\%) and d-separation (68.6\%) tasks
    \item \textbf{Robustness}: Adversarial tests confirm structural learning (67-70\%) vs. catastrophic failure (43-71\%)
\end{enumerate}

\textbf{Practical impact}: Semantic loss transforms causal reasoning fine-tuning from fundamentally broken (100\% collapse) to reliably stable. This is not an optimization—it is essential for any practical deployment.

\textbf{Broader implications}: Our findings suggest that fine-tuning on complex reasoning tasks may require task-specific inductive biases beyond standard cross-entropy loss. Future work on mathematical reasoning, logical inference, and other structured tasks should carefully monitor for similar collapse phenomena.

\bibliographystyle{plain}

\appendix

\section{Implementation Details}

\subsection{Data Generation Pipeline}

We implement a comprehensive synthetic data generation framework for causal reasoning tasks, consisting of two primary modules: a base generator for standard training and evaluation data, and a specialized adversarial generator for robustness testing.

\subsubsection{Graph Generation Algorithms}

Our framework employs two distinct graph generation strategies based on task requirements:

\paragraph{Sequential Chain Generation} For transitivity reasoning tasks, we generate directed acyclic chains of length $\ell$ where nodes $V = \{v_1, v_2, \ldots, v_\ell\}$ are connected by edges $E = \{(v_i, v_{i+1}) \mid i \in [1, \ell-1]\}$. To introduce structural variation, we apply edge flipping with probability $p_{\text{flip}} \in \{0.0, 0.3, 0.5\}$, reversing the direction of individual edges while maintaining overall connectivity.

Node names are randomly generated strings of length $n \sim \mathcal{U}(n_{\min}, n_{\max})$ from the alphabet $\Sigma = \{a\text{-}z, A\text{-}Z, 0\text{-}9\}$, where:
\begin{itemize}
    \item Training distribution: $n \in [1, 3]$
    \item Evaluation distribution: $n \in [8, 10]$ (for name length generalization)
\end{itemize}

\paragraph{DAG Generation with Controlled Branching} For d-separation tasks requiring more complex graph structures, we implement a topologically-ordered DAG generator. Given parameters $(|V|, \rho)$ where $\rho$ is edge density:

\begin{algorithm}[H]
\caption{Controlled DAG Generation}
\begin{algorithmic}[1]
\STATE Initialize nodes $V = \{v_1, \ldots, v_{|V|}\}$ with random names
\STATE $E \leftarrow \emptyset$
\FOR{$i = 1$ to $|V|$}
    \STATE $k \leftarrow \min(\lfloor |V| \cdot \rho \rfloor, 5)$ 
    \STATE $T \leftarrow$ sample $k$ nodes from $\{v_{i+1}, \ldots, v_{|V|}\}$
    \STATE $E \leftarrow E \cup \{(v_i, v_j) \mid v_j \in T\}$
\ENDFOR
\IF{$|E| < |V| - 1$}
\STATE \textit{Add backbone chain edges}
\ENDIF
\STATE \textbf{return} $G = (V, E)$
\end{algorithmic}
\end{algorithm}

Edge density ranges are task-specific:
\begin{itemize}
    \item Training: $\rho \sim \mathcal{U}(0.3, 0.6)$
    \item Evaluation: $\rho \sim \mathcal{U}(0.7, 1.2)$ (for branching complexity)
\end{itemize}

\subsubsection{Natural Language Template Generation}

Graphs are converted to natural language premises using deterministic templates:
\begin{verbatim}
premise: 
" ".join(
            [
                f"{a} causes {b}." 
                for (a,b) in E
            ]
        )
\end{verbatim}

For transitivity tasks, hypotheses query direct or transitive causation:
\begin{verbatim}
hypothesis: "Does {v_i} cause {v_j}?"
\end{verbatim}

For d-separation tasks, hypotheses include optional conditioning sets $Z \subset V$:
\begin{verbatim}
hypothesis:
"Are {v_i} and {v_j} d-separated given 
{Z}?"
\end{verbatim}
where $|Z| \leq 3$ is sampled uniformly.

\subsection{Causal Reasoning Algorithms}

\subsubsection{Transitivity Label Generation}

Labels are computed via depth-first search (DFS) for directed path existence:
\begin{algorithm}[H]
\caption{Path Existence Check: \textsc{FindPath}$(E, v_{\text{start}}, v_{\text{end}})$}
\begin{algorithmic}[1]
\IF{$v_{\text{start}} = v_{\text{end}}$} \RETURN \texttt{True} \ENDIF
\STATE $\text{visited} \leftarrow \emptyset$
\STATE $\text{stack} \leftarrow [v_{\text{start}}]$
\WHILE{$\text{stack} \neq \emptyset$}
    \STATE $v \leftarrow \text{stack.pop()}$
    \IF{$v = v_{\text{end}}$} \RETURN \texttt{True} \ENDIF
    \IF{$v \in \text{visited}$} \STATE \textbf{continue} \ENDIF
    \STATE $\text{visited} \leftarrow \text{visited} \cup \{v\}$
    \STATE $\text{stack.extend}(\{u \mid (v, u) \in E\})$
\ENDWHILE
\RETURN \texttt{False}
\end{algorithmic}
\end{algorithm}

\subsubsection{D-separation Algorithm}

We implement Pearl's d-separation criterion \cite{pearl2009causality} to determine conditional independence. The algorithm:

\begin{enumerate}
    \item \textbf{Path Finding:} Identify all undirected paths $\mathcal{P}(v_i, v_j)$ between query nodes using breadth-first search with path length limit $L_{\max} = 10$.
    
    \item \textbf{Blocking Rule Evaluation:} For each path $p = (v_i, \ldots, v_j) \in \mathcal{P}(v_i, v_j)$ and each intermediate node $v_k$ with neighbors $(v_{k-1}, v_{k+1})$:
    
    \begin{itemize}
        \item \textbf{Collider:} If $(v_{k-1}, v_k) \in E$ and $(v_{k+1}, v_k) \in E$, path is blocked unless $v_k \in Z$ or $\exists v_d \in \text{descendants}(v_k): v_d \in Z$
        \item \textbf{Chain:} If $(v_{k-1}, v_k) \in E$ and $(v_k, v_{k+1}) \in E$, path is blocked if $v_k \in Z$
        \item \textbf{Fork:} If $(v_k, v_{k-1}) \in E$ and $(v_k, v_{k+1}) \in E$, path is blocked if $v_k \in Z$
    \end{itemize}
    
    \item \textbf{D-separation Decision:} Return \texttt{True} if all paths are blocked, \texttt{False} otherwise.
\end{enumerate}

To handle descendant queries efficiently, we implement a memoized BFS traversal with visited set tracking.

\subsection{Multi-Stage Validation Framework}

Each generated example undergoes rigorous validation to ensure logical consistency:

\paragraph{Premise Parsing} Causal edges are extracted using regex pattern matching:
\begin{verbatim}
pattern: r"(\w+) causes (\w+)"
\end{verbatim}
Failed parses are rejected (acceptance rate: $>99\%$).

\paragraph{Hypothesis Parsing} Query nodes and conditioning sets are extracted via string decomposition with error handling for malformed queries.

\paragraph{Label Verification} Ground truth labels are recomputed from parsed graphs and compared against generated labels. Examples with mismatches are rejected.

\paragraph{Graph Validity Checks} For d-separation tasks, we reject graphs with:
\begin{itemize}
    \item $|E| < |V| - 1$ (insufficient connectivity)
    \item $|E| > 3|V|$ (excessive density)
    \item Unreachable node pairs with empty path sets
\end{itemize}

\subsection{Optimization Strategies}

\subsubsection{Computational Optimizations}

\begin{itemize}
    \item \textbf{Memoization:} D-separation checks are cached using LRU cache with 1000-entry limit, reducing redundant path computations for isomorphic subgraphs.
    
    \item \textbf{Early Rejection:} Invalid graphs are filtered before expensive d-separation computation based on structural heuristics (edge count bounds, node reachability).
    
    \item \textbf{Attempt Limits:} Generation retries are capped at 10 attempts per example to prevent infinite loops on infeasible configurations.
    
    \item \textbf{Path Length Limits:} BFS path finding terminates at depth 10, trading completeness for tractability on large graphs.
\end{itemize}

\subsubsection{Acceptance Rate Analysis}

Table \ref{tab:acceptance} shows generation acceptance rates across tasks:

\begin{table}[h]
\centering
\small
\caption{Generation Acceptance Rates}
\label{tab:acceptance}
\begin{tabular}{lcc}
\toprule
\textbf{Task} & \textbf{Target} & \textbf{Acceptance Rate} \\
\midrule
Transitivity (Train) & 50,000 & $>95\%$ \\
D-separation (Train) & 50,000 & $\sim 70\%$ \\
Branching (Eval) & 10,000 & $\sim 40\%$ \\
Adversarial (Eval) & 1,000 & $\sim 65\%$ \\
\bottomrule
\end{tabular}
\end{table}

Lower acceptance for d-separation reflects stricter validation requirements and graph complexity constraints.

\subsection{Dataset Composition}

\subsubsection{Training Datasets}

\begin{itemize}
    \item \textbf{Transitivity Training} (\texttt{transitivity\_train.jsonl}): 50,000 examples
    \begin{itemize}
        \item Chain length: $\ell \sim \mathcal{U}(3, 6)$
        \item Node names: $n \sim \mathcal{U}(1, 3)$
        \item Edge flipping: $p \in \{0.0, 0.3, 0.5\}$
    \end{itemize}
    
    \item \textbf{D-separation Training} (\texttt{dsep\_train.jsonl}): 50,000 examples
    \begin{itemize}
        \item Graph size: $|V| \sim \mathcal{U}(3, 6)$
        \item Edge density: $\rho \sim \mathcal{U}(0.3, 0.6)$
        \item Conditioning set size: $|Z| \sim \mathcal{U}(0, 3)$
    \end{itemize}
\end{itemize}

\subsubsection{Standard Evaluation Datasets}

Five evaluation sets test different generalization capabilities (10,000 examples each):

\begin{enumerate}
    \item \textbf{Length Generalization} (\path{length_eval.jsonl}): Chain length $\ell \sim \mathcal{U}(7, 15)$
    
    \item \textbf{Structural Variation} (\path{reversed_eval.jsonl}): All edges reversed, $E' = \{(b, a) \mid (a, b) \in E\}$
    
    \item \textbf{Order Invariance} (\path{shuffled_eval.jsonl}): Premise statements randomly permuted with $p_{\text{flip}} = 0.5$
    
    \item \textbf{Name Length Generalization} (\path{long_names_eval.jsonl}): Node names $n \sim \mathcal{U}(8, 10)$
    
    \item \textbf{Branching Complexity} (\path{branching_eval.jsonl}): DAGs with $\rho \sim \mathcal{U}(0.7, 1.2)$ and $\ell \sim \mathcal{U}(7, 15)$
\end{enumerate}

\subsubsection{Adversarial Evaluation Dataset}

The adversarial evaluation set (\path{adversarial_eval.jsonl}, 1,000 examples) targets specific failure modes through carefully designed graph construction strategies:

\paragraph{Irrelevant Nodes (30\%)} These examples test whether models can focus on relevant causal structure while ignoring disconnected components. Generation procedure:
\begin{enumerate}
    \item Generate a main chain of length $\ell_{\text{main}} \sim \mathcal{U}(3, 5)$ with standard parameters
    \item Add $k \sim \mathcal{U}(1, 3)$ disconnected chains, each of length $\ell_{\text{irrel}} \sim \mathcal{U}(2, 4)$
    \item Ensure node name uniqueness across all chains through rejection sampling (maximum 10 attempts)
    \item Query exclusively about nodes within the main chain: $v_i, v_j \in V_{\text{main}}$
    \item Premise contains edges from all chains: $E = E_{\text{main}} \cup E_{\text{irrel},1} \cup \ldots \cup E_{\text{irrel},k}$
\end{enumerate}

Example structure:

{\small
\begin{verbatim}
Premise: "A causes B. B causes C. 
X causes Y. P causes Q. Q causes R."
[main]  [---irrelevant chains---]

Hypothesis: "Does A cause C?"
Label: "Yes"
\end{verbatim}
}

This tests whether models erroneously incorporate irrelevant nodes into reasoning or correctly isolate the queried subgraph.

\paragraph{Broken Chains (30\%)} These examples test detection of non-existent causal paths across disconnected graph components. Generation procedure:
\begin{enumerate}
    \item Generate $k \sim \mathcal{U}(2, 3)$ completely disconnected chains
    \item Each chain has length $\ell_i \sim \mathcal{U}(2, 4)$
    \item Enforce strict node name disjointness: $V_i \cap V_j = \emptyset$ for $i \neq j$
    \item Query across different components: select $v_i \in V_a$ and $v_j \in V_b$ where $a \neq b$
    \item Label is always ``No'' since no path exists between disconnected components
\end{enumerate}

Example structure:

{\small
\begin{verbatim}
Premise: "A causes B. B causes C. 
X causes Y. P causes Q."
[chain 1] [chain 2] [chain 3]

Hypothesis: "Does A cause Y?"
Label: "No"
\end{verbatim}
}

This evaluates whether models incorrectly hallucinate transitive connections across graph boundaries or properly recognize component isolation.

\paragraph{Extended Transitivity (40\%)} These examples test multi-hop reasoning beyond the training distribution length. Generation procedure:
\begin{enumerate}
    \item Generate sequential chains with $\ell \sim \mathcal{U}(7, 12)$, exceeding training maximum of 6
    \item Use standard edge generation without flipping: $E = \{(v_i, v_{i+1}) \mid i \in [1, \ell-1]\}$
    \item Query endpoint causation: ``Does $v_1$ cause $v_\ell$?''
    \item Label is always ``Yes'' requiring $\ell - 1$ transitive steps
\end{enumerate}

Example structure:

{\small
\begin{verbatim}
Premise: "A causes B. B causes C. 
C causes D. D causes E. E causes F. 
F causes G. G causes H. H causes I. 
I causes J."
[9-hop chain, exceeds training max]

Hypothesis: "Does A cause J?"
Label: "Yes"
\end{verbatim}
}

This probes compositional generalization: whether models can chain reasoning beyond training-time depth limits.

\paragraph{Validation and Quality Control} All adversarial examples undergo identical validation as training data:
\begin{itemize}
    \item Premise parsing verification (regex extraction of all edges)
    \item Label recomputation using \textsc{FindPath} algorithm
    \item Graph connectivity checks (appropriate for disconnected graph examples)
    \item Maximum 15 generation attempts per example (higher than standard 10 due to structural constraints)
\end{itemize}

Acceptance rates vary by adversarial type: irrelevant nodes ($\sim$70\%), broken chains ($\sim$60\%), extended transitivity ($\sim$65\%), yielding overall acceptance of $\sim$65\% for the adversarial set.

\subsection{Implementation Details}

The complete data generation pipeline is implemented in Python 3.8+ using:
\begin{itemize}
    \item \texttt{dataclasses} for configuration management
    \item \texttt{collections.deque} for efficient BFS implementation
    \item \texttt{functools.lru\_cache} for memoization
    \item \texttt{logging} for progress tracking and debugging
\end{itemize}

All datasets are serialized as JSONL files with schema:

{\small
\begin{verbatim}
{
  "premise": str,    # Causal statements
  "hypothesis": str, # Query question
  "label": str       # "Yes" or "No"
}
\end{verbatim}
}

Generation time averages 0.02s per transitivity example and 0.15s per d-separation example on a single CPU core, with total pipeline runtime under 3 hours for all 121,000 examples.

\subsection{Hardware and Runtime}

All experiments conducted on Google Colab Pro with NVIDIA T4 GPU (16GB). Training time per model:
\begin{itemize}
    \item Baseline (no semantic): ~45 minutes
    \item Semantic loss: ~52 minutes (+15\% overhead)
\end{itemize}

\section{Additional Experimental Results}

\subsection{Per-Task Confusion Matrices}

This section provides detailed confusion matrices for all model variants across standard and adversarial evaluation sets. Confusion matrices reveal the true nature of model predictions beyond aggregate accuracy metrics, particularly exposing collapse patterns through extreme TP/FP or TN/FN distributions.

\subsubsection{Standard Evaluation (10,000 samples per task)}

Tables \ref{tab:cm_standard_gemma} through \ref{tab:cm_dsep_sem_v2} show confusion matrices across the five standard generalization tests. Key patterns:

\begin{itemize}
\item \textbf{Transitivity V1 collapse:} TP = 10,000 (length task), TN = 0 across all tasks $\rightarrow$ always predicts "Yes"
\item \textbf{D-separation V1 collapse:} TP near-zero, TN dominates $\rightarrow$ always predicts "No"  
\item \textbf{Semantic models:} Balanced TP/TN/FP/FN distributions indicate context-dependent predictions
\end{itemize}

\begin{table*}[!htbp]
\centering
\small
\caption{Standard Gemma: Confusion matrices across standard evaluation tasks}
\label{tab:cm_standard_gemma}
\begin{tabular}{lccccc}
\hline
\textbf{Metric} & \textbf{Length} & \textbf{Branch} & \textbf{Rev} & \textbf{Shuff} & \textbf{LongN} \\
\hline
True Positive & 5716 & 0 & 80 & 8 & 1202 \\
True Negative & 0 & 9792 & 5938 & 7058 & 5260 \\
False Positive & 0 & 12 & 3836 & 2927 & 1304 \\
False Negative & 4284 & 196 & 146 & 7 & 2234 \\
\hline
\end{tabular}
\end{table*}

\begin{table*}[!htbp]
\centering
\small
\caption{Transitivity V1 (Collapsed): Confusion matrices showing complete collapse to "Yes" predictions}
\label{tab:cm_transitivity_v1}
\begin{tabular}{lccccc}
\hline
\textbf{Metric} & \textbf{Length} & \textbf{Branch} & \textbf{Rev} & \textbf{Shuff} & \textbf{LongN} \\
\hline
True Positive & 10000 & 196 & 226 & 15 & 3436 \\
True Negative & 0 & 0 & 0 & 0 & 0 \\
False Positive & 0 & 9804 & 9774 & 9985 & 6564 \\
False Negative & 0 & 0 & 0 & 0 & 0 \\
\hline
\end{tabular}
\end{table*}

\begin{table*}[!htbp]
\centering
\small
\caption{D-separation V1 (Collapsed): Confusion matrices showing collapse to "No" predictions}
\label{tab:cm_dseparation_v1}
\begin{tabular}{lccccc}
\hline
\textbf{Metric} & \textbf{Length} & \textbf{Branch} & \textbf{Rev} & \textbf{Shuff} & \textbf{LongN} \\
\hline
True Positive & 1889 & 0 & 17 & 4 & 12 \\
True Negative & 0 & 9804 & 9006 & 9639 & 6564 \\
False Positive & 0 & 0 & 768 & 346 & 0 \\
False Negative & 8111 & 196 & 209 & 11 & 3424 \\
\hline
\end{tabular}
\end{table*}

\begin{table*}[!htbp]
\centering
\small
\caption{Transitivity Semantic V4: Confusion matrices showing balanced predictions}
\label{tab:cm_transitivity_sem_v4}
\begin{tabular}{lccccc}
\hline
\textbf{Metric} & \textbf{Length} & \textbf{Branch} & \textbf{Rev} & \textbf{Shuff} & \textbf{LongN} \\
\hline
True Positive & 6464 & 1 & 89 & 9 & 1864 \\
True Negative & 0 & 9788 & 5599 & 6962 & 4411 \\
False Positive & 0 & 16 & 4175 & 3023 & 2153 \\
False Negative & 3536 & 195 & 137 & 6 & 1572 \\
\hline
\end{tabular}
\end{table*}

\begin{table*}[!htbp]
\centering
\small
\caption{D-separation Semantic V2: Confusion matrices showing balanced predictions}
\label{tab:cm_dsep_sem_v2}
\begin{tabular}{lccccc}
\hline
\textbf{Metric} & \textbf{Length} & \textbf{Branch} & \textbf{Rev} & \textbf{Shuff} & \textbf{LongN} \\
\hline
True Positive & 6283 & 0 & 91 & 8 & 1405 \\
True Negative & 0 & 9777 & 5318 & 6487 & 4952 \\
False Positive & 0 & 27 & 4456 & 3498 & 1612 \\
False Negative & 3717 & 196 & 135 & 7 & 2031 \\
\hline
\end{tabular}
\end{table*}

\subsubsection{Adversarial Evaluation (1,000 samples)}

Table \ref{tab:cm_adversarial} shows confusion matrices on the adversarial structural robustness test. This evaluation distinguishes structural understanding from heuristics through challenging examples with irrelevant nodes, broken chains, and extended transitivity.

\textbf{Key findings:}
\begin{itemize}
\item \textbf{Transitivity V1:} TP = 708, TN = 0, FN = 0 $\rightarrow$ Maintains collapse even on adversarial distribution
\item \textbf{D-separation V1:} TP = 247, FN = 461 $\rightarrow$ Catastrophic recall failure (34.9\%)
\item \textbf{Semantic models:} Balanced confusion matrices with TP/TN/FP/FN all non-zero $\rightarrow$ Context-dependent reasoning
\end{itemize}

\begin{table*}[!htbp]
\centering
\small
\caption{Adversarial evaluation confusion matrices (1,000 samples testing structural understanding)}
\label{tab:cm_adversarial}
\begin{tabular}{lccccc}
\hline
\textbf{Model} & \textbf{TP} & \textbf{TN} & \textbf{FP} & \textbf{FN} & \textbf{Interpretation} \\
\hline
Standard Gemma & 533 & 134 & 158 & 175 & Heuristic-based \\
Transitivity V1 & 708 & 0 & 292 & 0 & Collapsed (always Yes) \\
D-separation V1 & 247 & 183 & 109 & 461 & Catastrophic failure \\
Transitivity Sem V4 & 588 & 110 & 182 & 120 & Structural reasoning \\
D-separation Sem V2 & 554 & 124 & 168 & 154 & Structural reasoning \\
\hline
\end{tabular}
\end{table*}

\subsubsection{Interpretation Guidelines}

The confusion matrices reveal three distinct behavioral patterns:

\textbf{1. Catastrophic Collapse (V1 models):}
\begin{itemize}
\item Transitivity V1: TN = 0 across all tasks, indicating exclusive "Yes" predictions
\item D-separation V1: TP near-zero with massive FN counts, indicating exclusive "No" predictions
\item These patterns are input-independent, confirming prediction bias collapse
\end{itemize}

\textbf{2. Heuristic-Based Predictions (Standard Gemma):}
\begin{itemize}
\item Task-specific patterns (e.g., 0\% branching accuracy = all "No")
\item Moderate TP/TN values with significant FP/FN errors
\item Performance varies dramatically by task type
\end{itemize}

\textbf{3. Structural Reasoning (Semantic models):}
\begin{itemize}
\item All four values (TP/TN/FP/FN) non-zero and substantial
\item TP and TN values proportional to label distributions
\item Consistent error patterns across tasks, not task-specific collapse
\end{itemize}

\textbf{Critical diagnostic insight:} Accuracy alone cannot detect collapse. For example, D-separation V1 achieves 73.9\% average accuracy (Table~\ref{tab:collapse_evidence}) while exhibiting severe FN bias (8,111 false negatives on length task). Only examination of the full confusion matrix reveals this catastrophic failure mode, highlighting the necessity of comprehensive metric reporting for causal reasoning evaluation.

\subsection{Prediction Distribution Histograms}

\subsubsection{Standard Evaluation}

\begin{figure}[!htbp]
    \centering
    \includegraphics[width=0.9\linewidth]{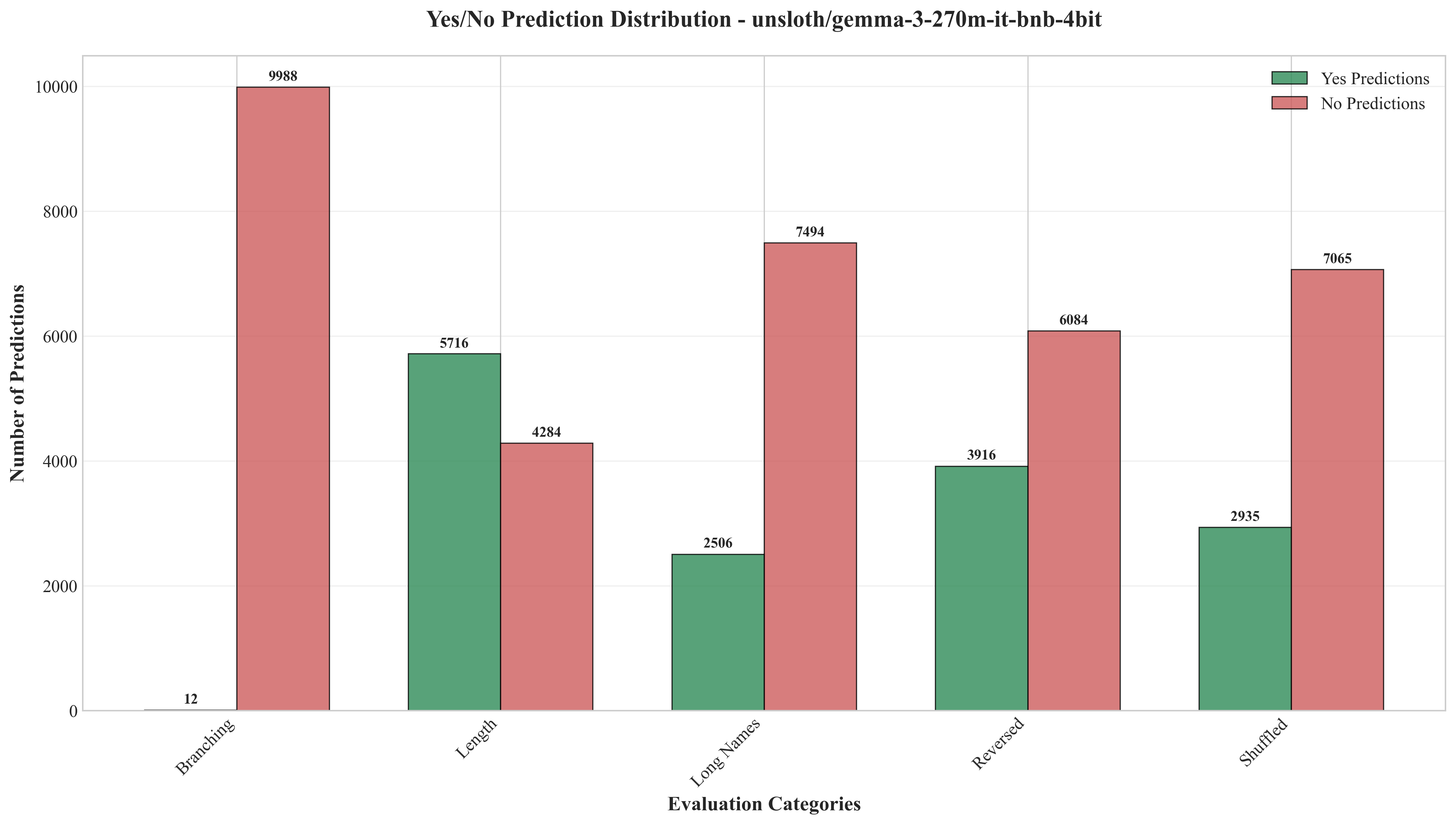}
    \caption{Prediction distribution of the pretrained Gemma-3 270M model on the standard evaluation suite (Length, Branching, Reversed, Shuffled, Long Names).}
    \label{fig:standard_original}
\end{figure}

\begin{figure}[!htbp]
    \centering
    \includegraphics[width=0.48\linewidth]{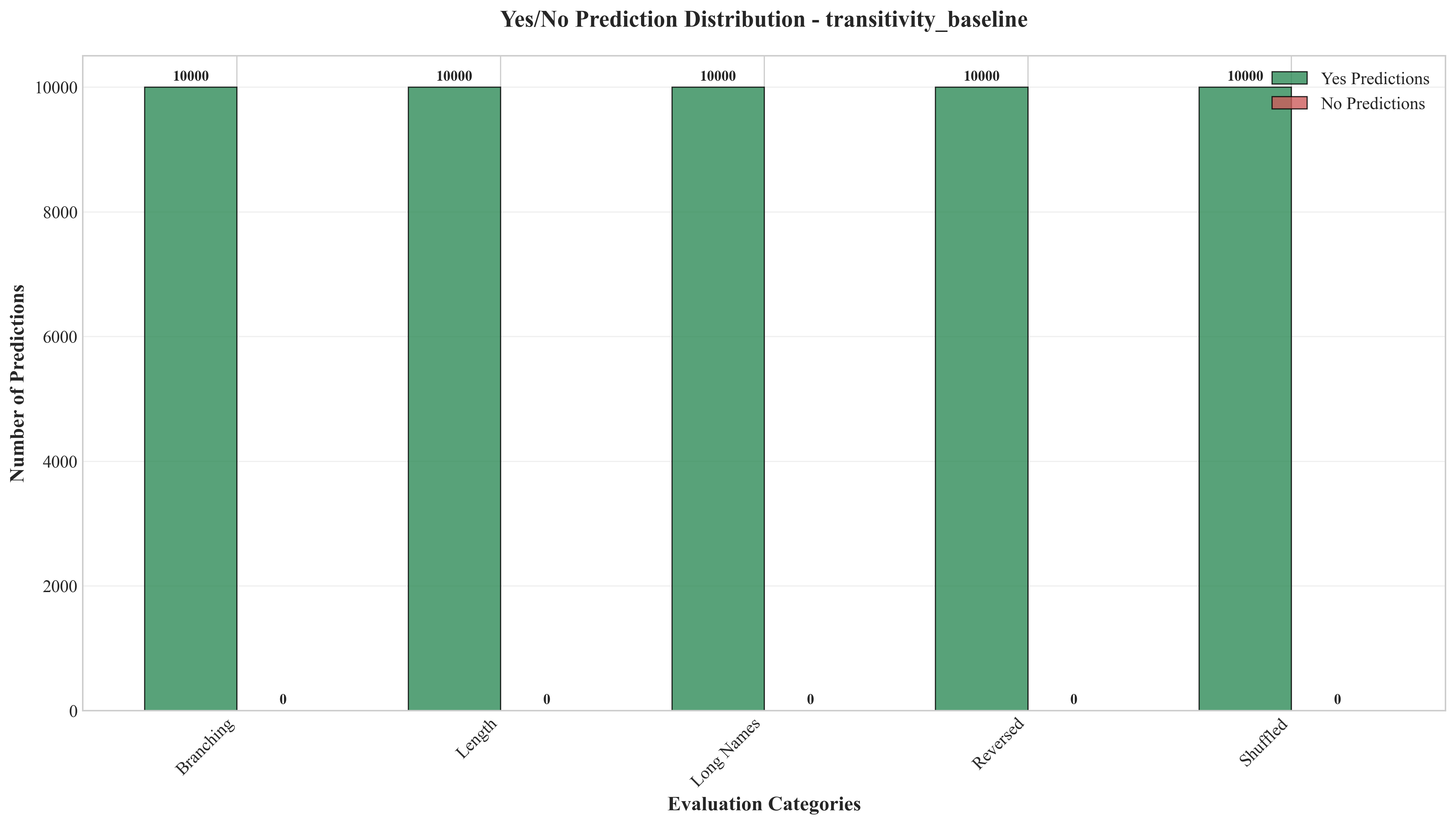}
    \includegraphics[width=0.48\linewidth]{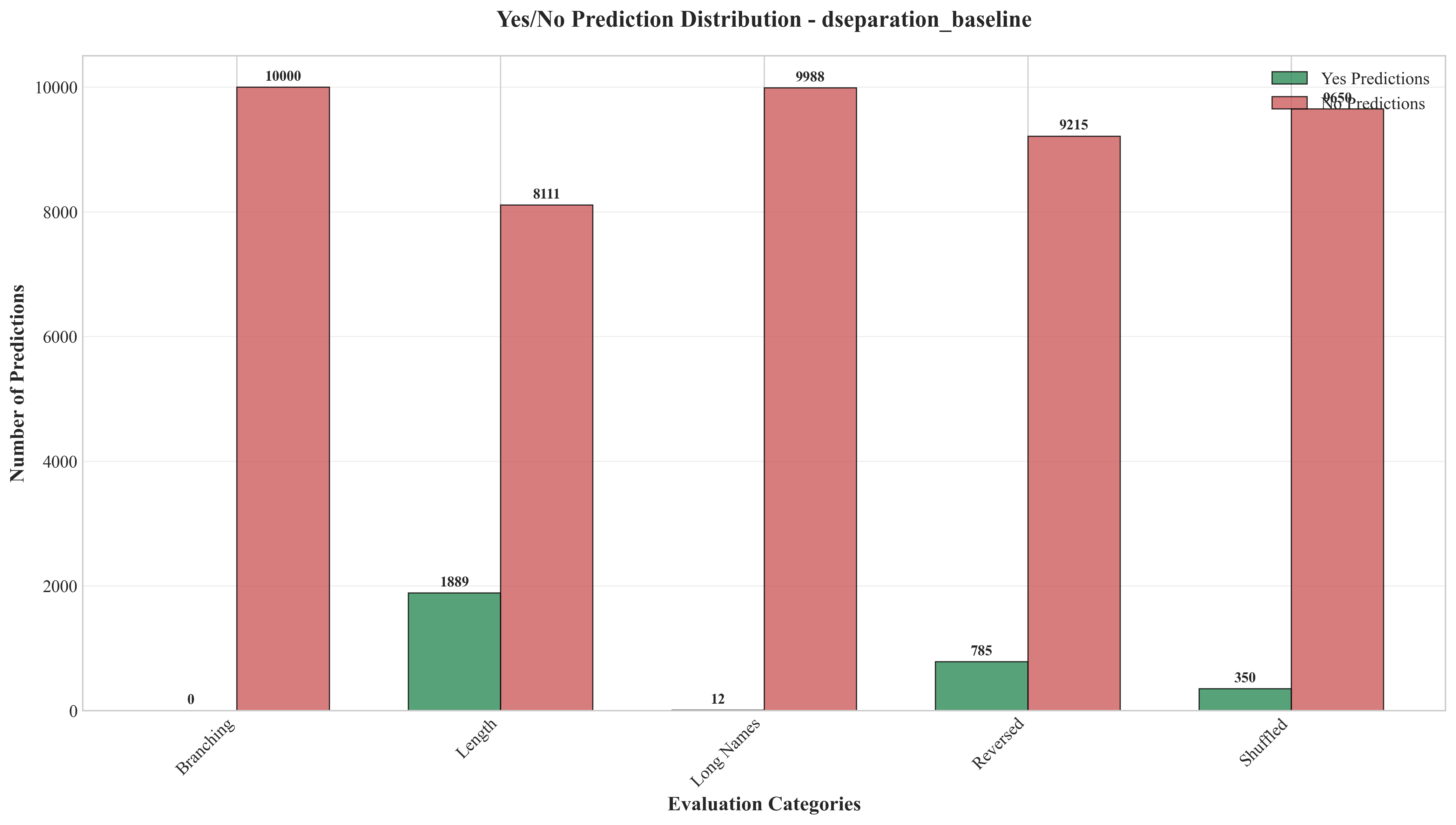}
    \caption{Collapsed baseline models: Transitivity V1 (left) and D-Separation V1 (right) on the standard evaluation tasks.}
    \label{fig:standard_baselines}
\end{figure}

\begin{figure}[!htbp]
    \centering
    \includegraphics[width=0.48\linewidth]{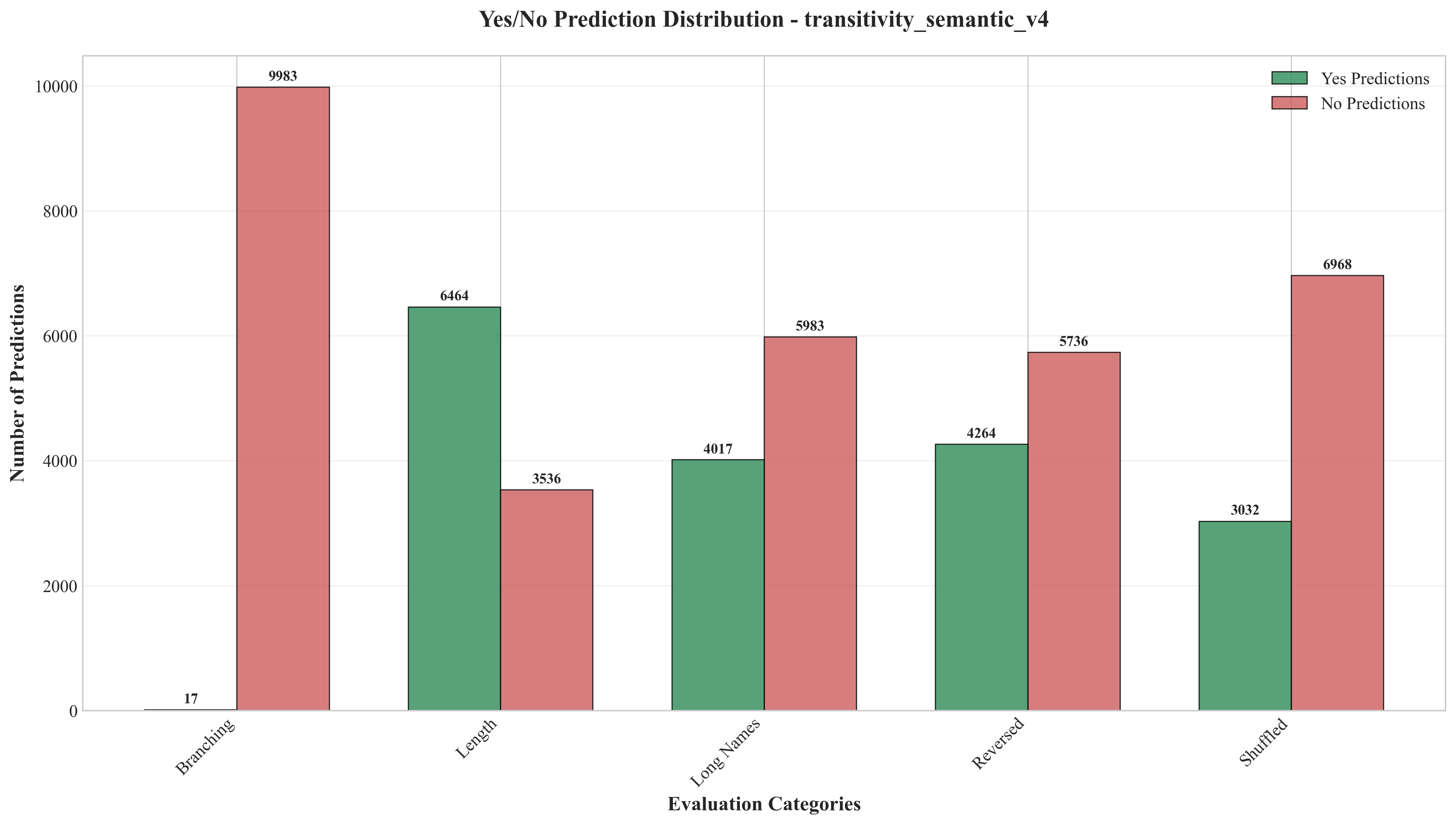}
    \includegraphics[width=0.48\linewidth]{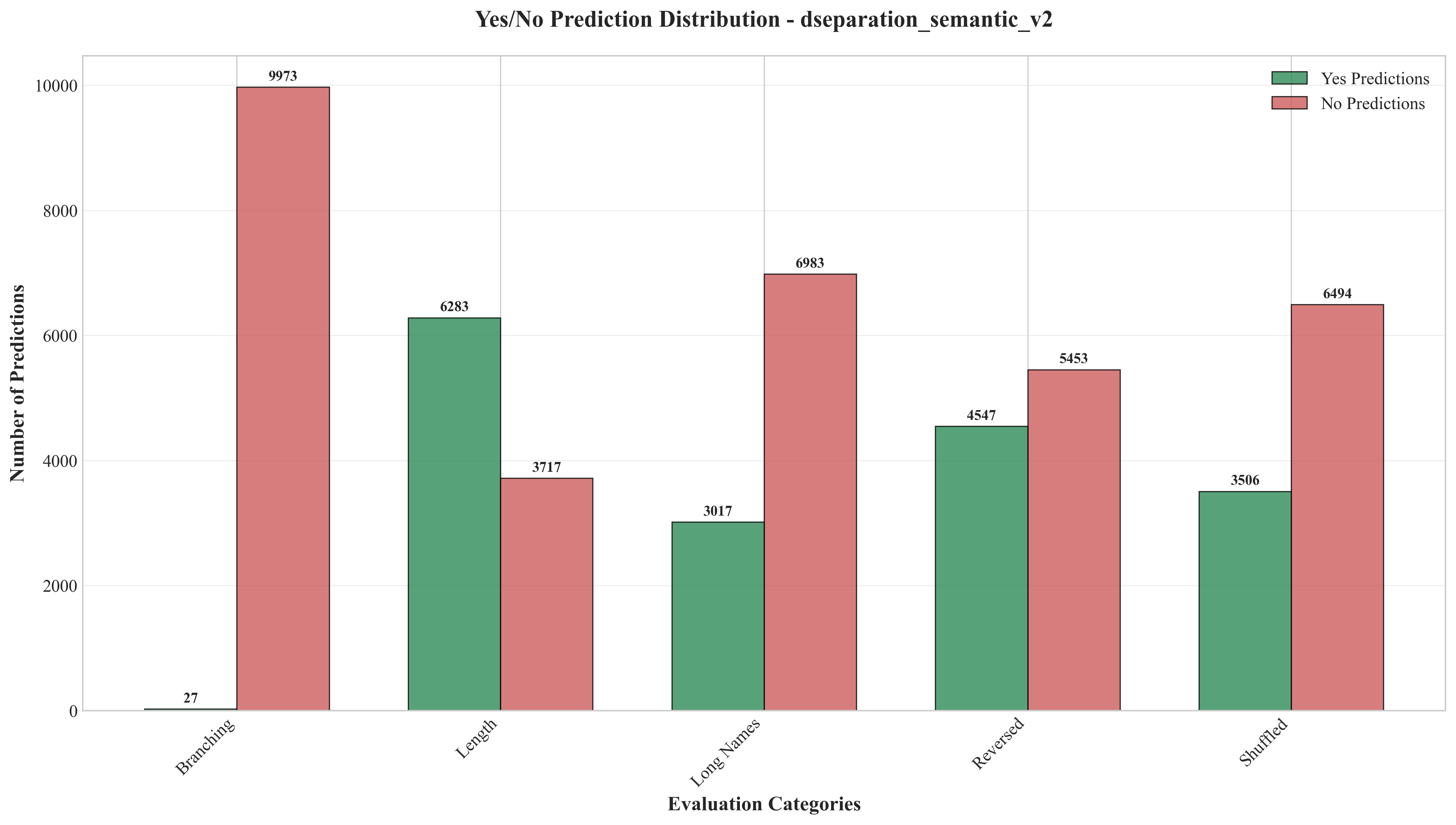}
    \caption{Semantic-loss fine-tuned models: Transitivity V4 (left) and D-Separation V2 (right) on the standard evaluation tasks.}
    \label{fig:standard_semantic}
\end{figure}

\FloatBarrier

\subsubsection{Adversarial Evaluation (Structural Robustness)}

\begin{figure}[!htbp]
    \centering
    \includegraphics[width=0.9\linewidth]{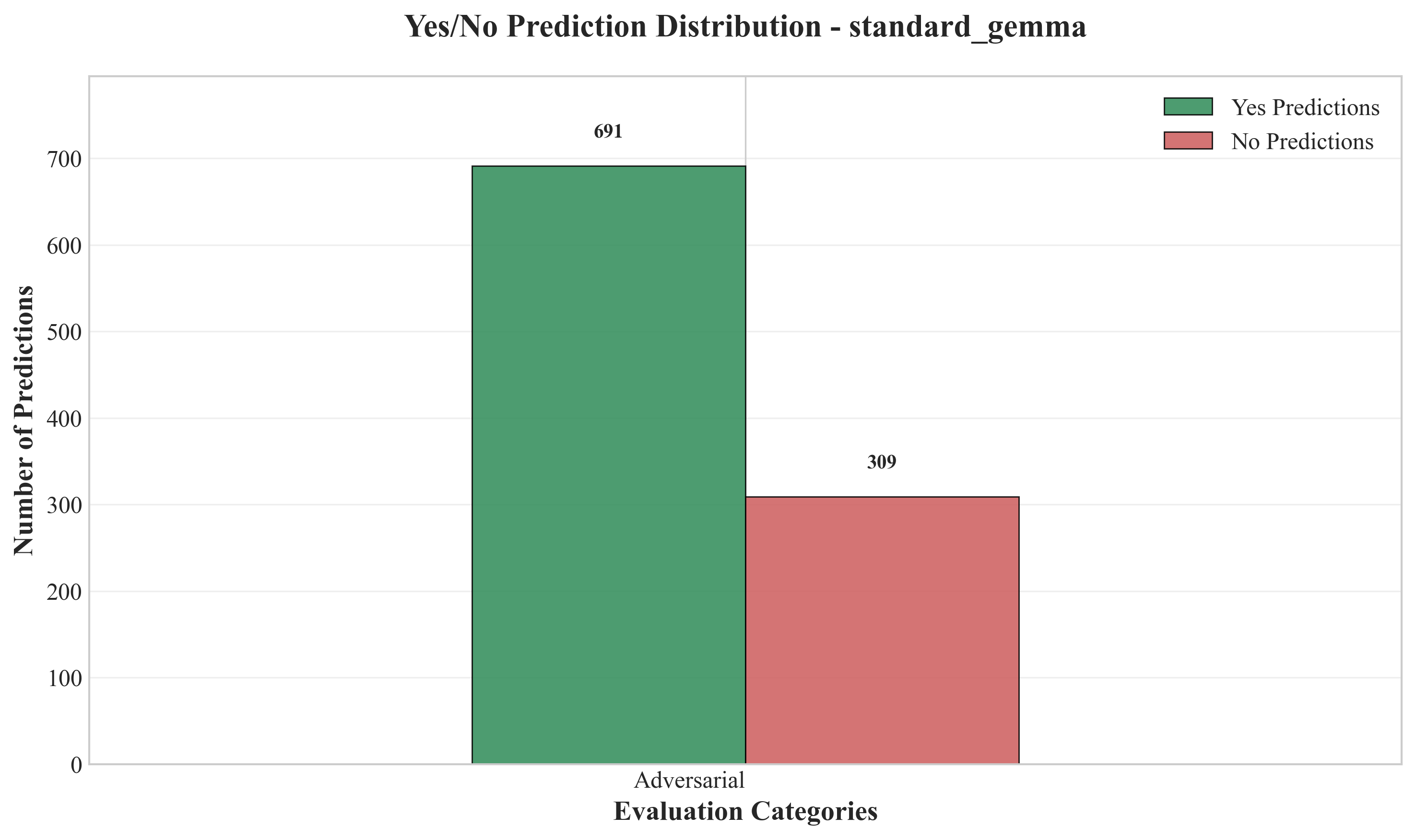}
    \caption{Pretrained Gemma-3 270M model on adversarial structural robustness tests (Irrelevant nodes, Broken chains, Long chains).}
    \label{fig:adv_original}
\end{figure}

\begin{figure}[!htbp]
    \centering
    \includegraphics[width=0.48\linewidth]{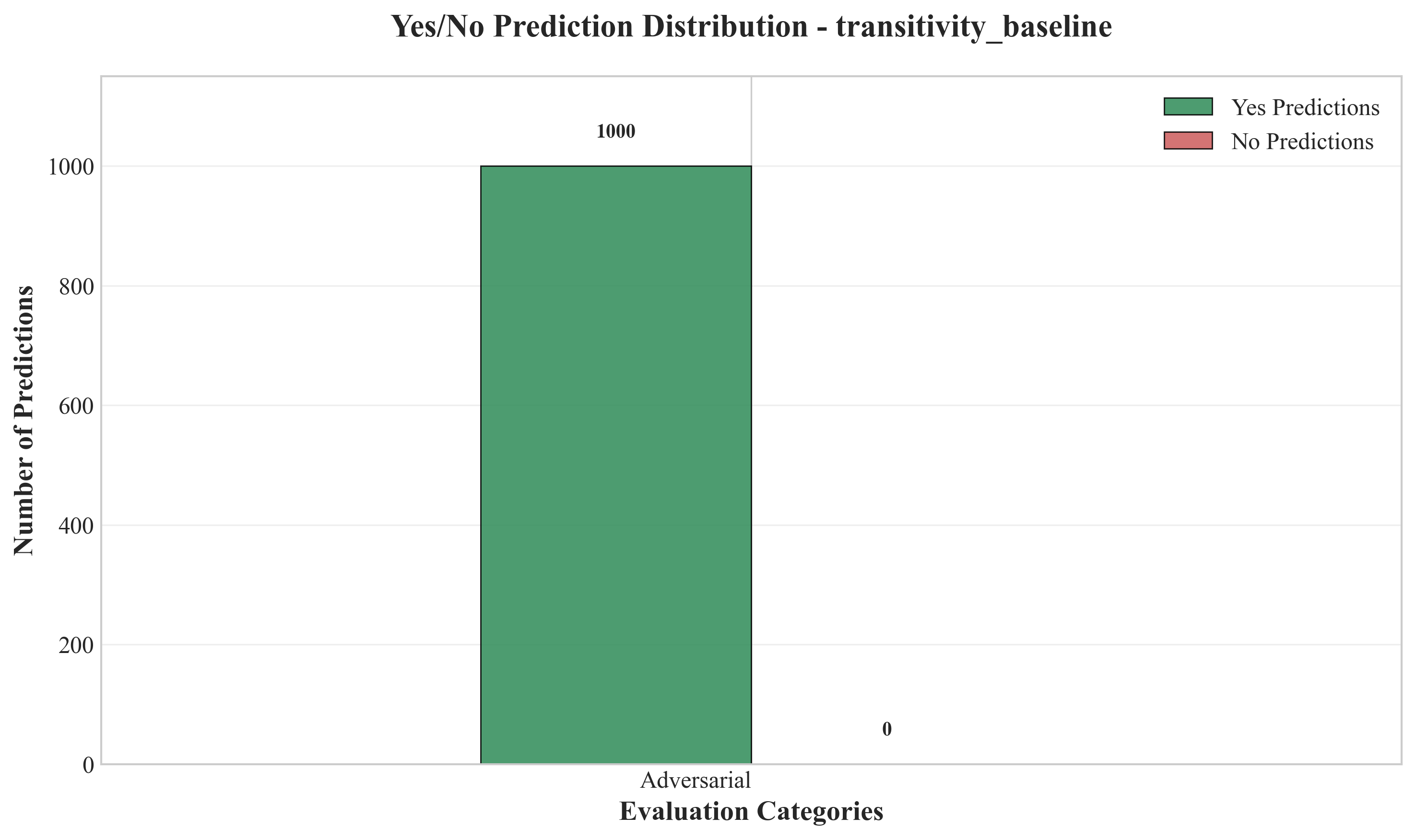}
    \includegraphics[width=0.48\linewidth]{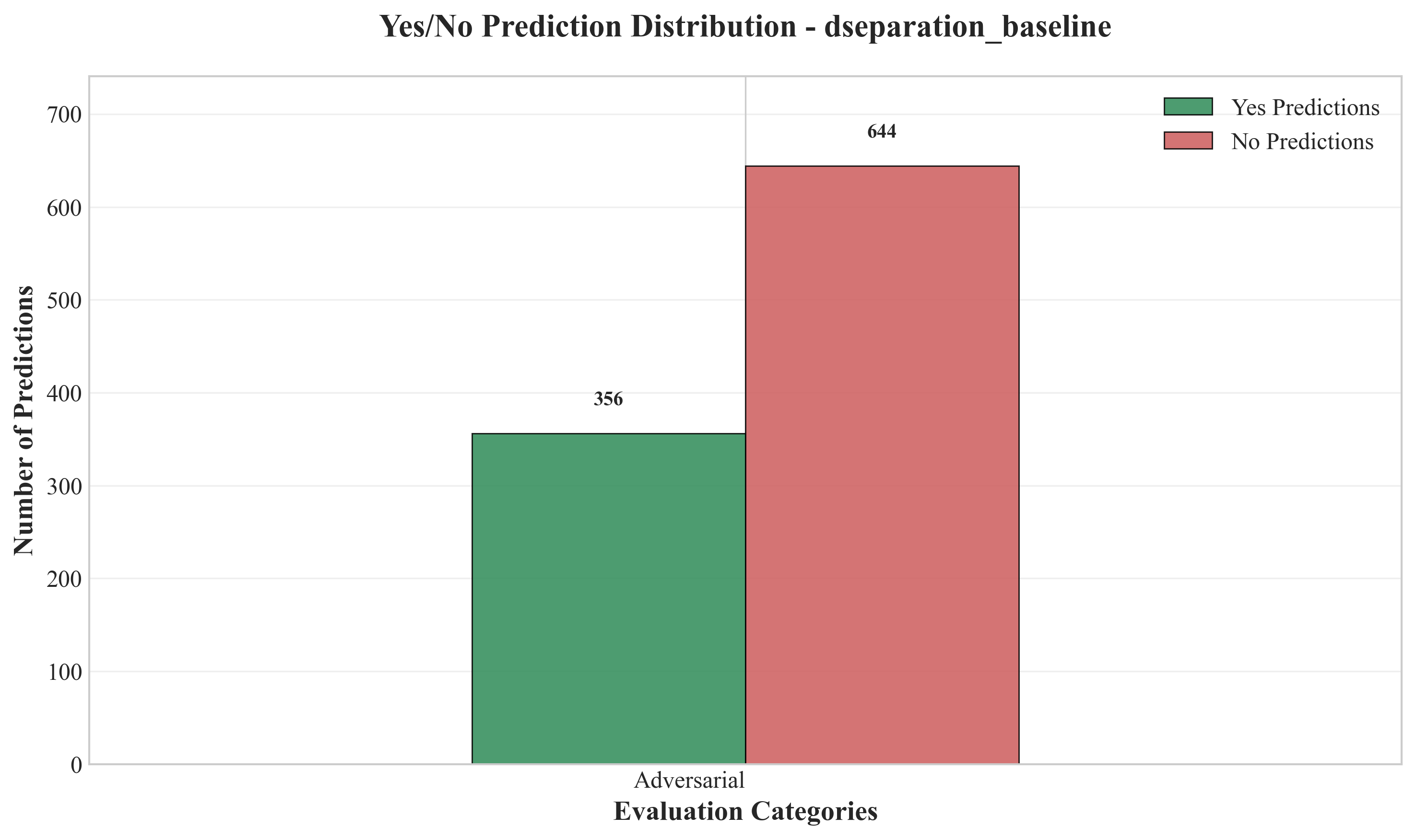}
    \caption{Collapsed baseline models (Transitivity V1, D-Separation V1) on adversarial examples.}
    \label{fig:adv_baselines}
\end{figure}
\clearpage
\begin{figure}[!htbp]
    \centering
    \includegraphics[width=0.48\linewidth]{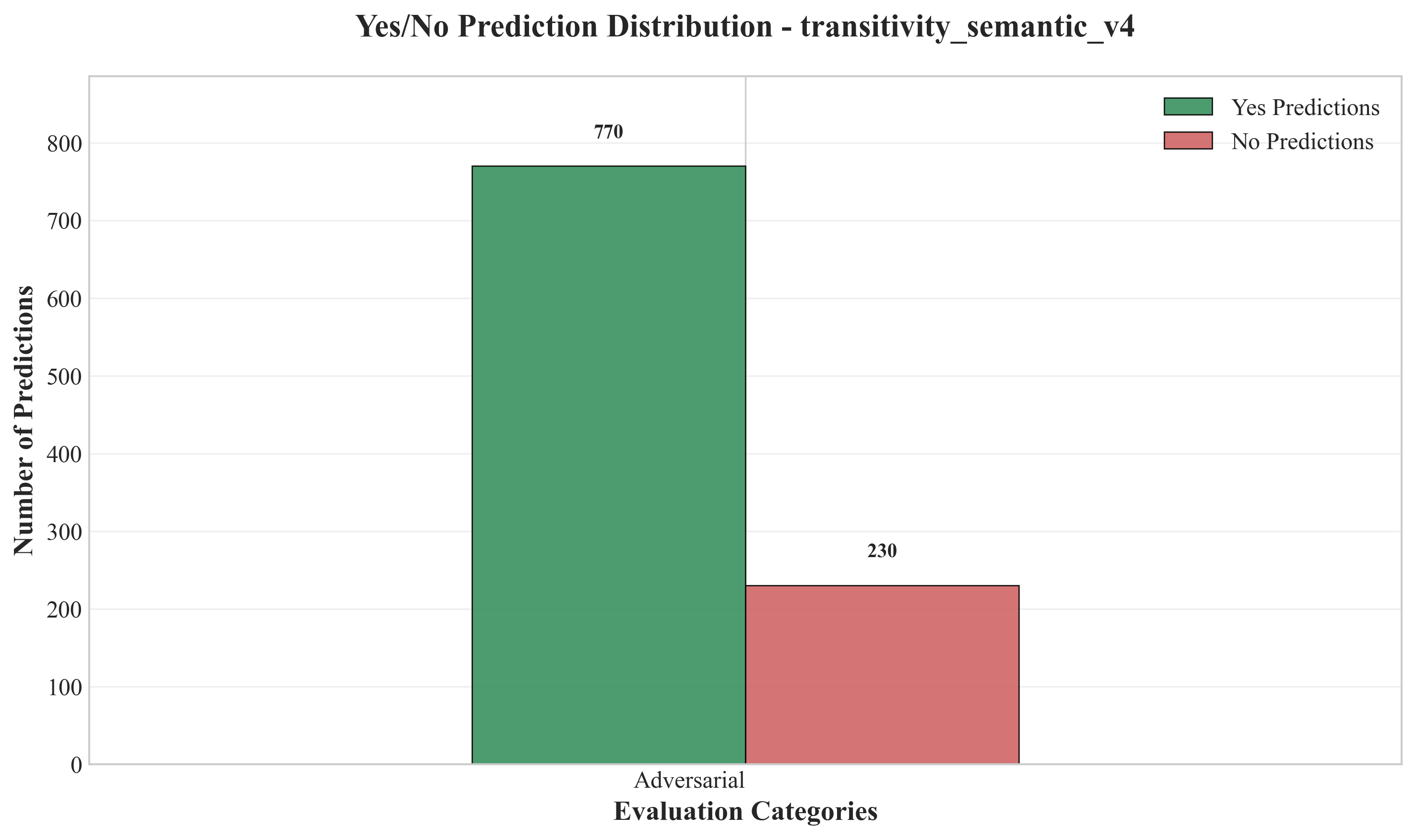}
    \includegraphics[width=0.48\linewidth]{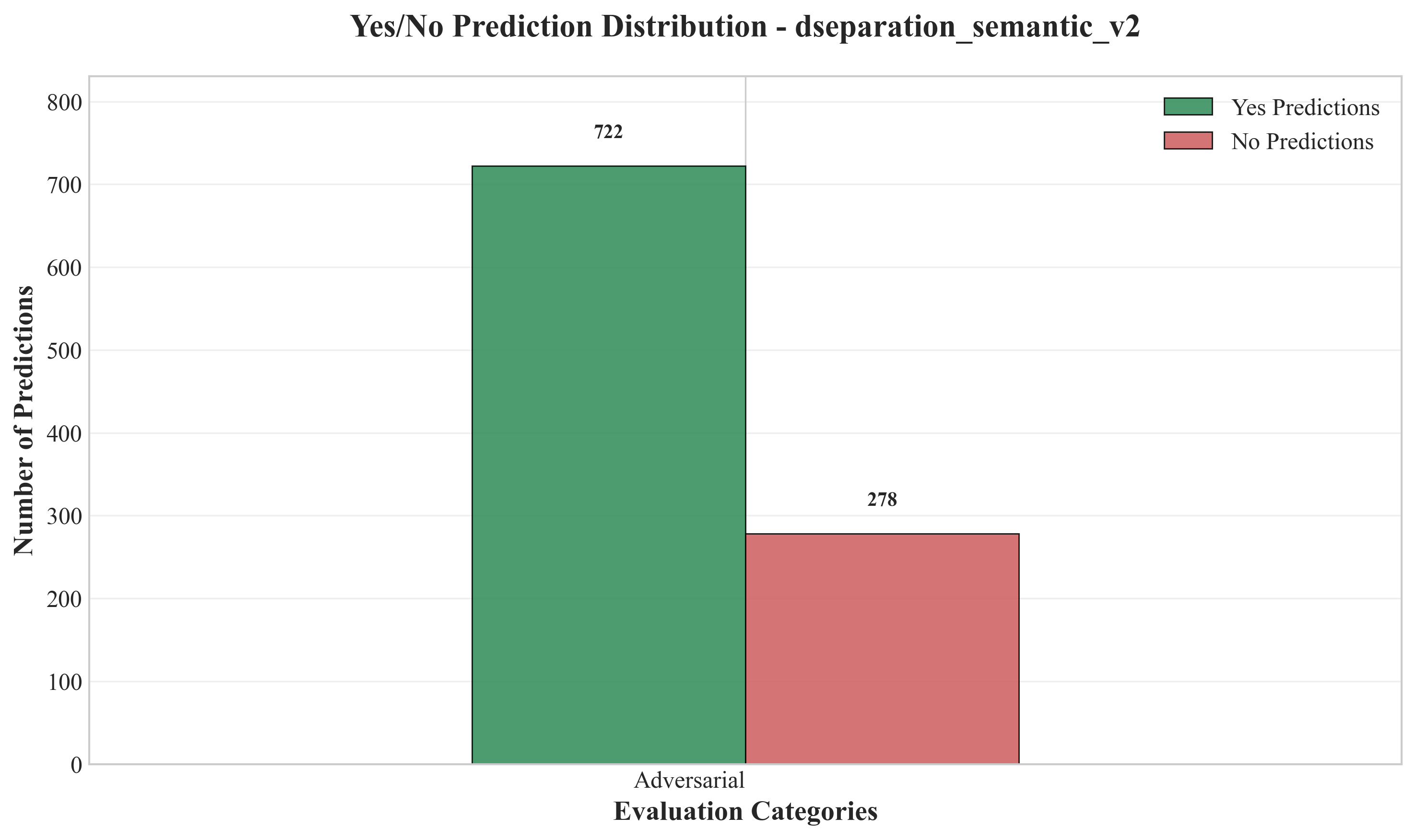}
    \caption{Semantic-loss fine-tuned models (Transitivity V4, D-Separation V2) on adversarial structural robustness tests.}
    \label{fig:adv_semantic}
\end{figure}

\section*{Code and Data Availability}

All code, trained models, and evaluation datasets are publicly available to ensure full reproducibility.

\begin{itemize}
    \item \textbf{Code \& Experiments:} The GitHub repository contains data generation scripts for both transitivity and d-separation tasks (generating the 50,000-sample training sets and adversarial evaluation sets), along with the comprehensive Colab notebook (\texttt{gemma\_semantic.ipynb}) documenting all experiments -- baseline fine-tuning, semantic loss versions V1 through V4, dynamic lambda scheduling implementation, and the full evaluation pipeline:\\
    \url{https://github.com/inquisitour/semantic-loss-causal-reasoning}

    \item \textbf{Trained Models:} All four model variants are hosted on Hugging Face (MIT license):\\
    \url{https://huggingface.co/ludwigw/gemma-transitivity-semantic-v4}\\
    \url{https://huggingface.co/ludwigw/gemma-dseparation-semantic-v2}\\
    \url{https://huggingface.co/ludwigw/gemma-transitivity-baseline}\\
    \url{https://huggingface.co/ludwigw/gemma-dseparation-baseline}

    \item \textbf{Datasets:} Training sets (50,000 examples per task), five standard evaluation sets (10,000 samples each: length, branching, reversed, shuffled, long names), and the adversarial structural robustness set (1,000 samples) are available at:\\
    \url{https://huggingface.co/datasets/ludwigw/causal-reasoning-benchmarks}
\end{itemize}

\end{document}